\documentclass[letterpaper, 10 pt, conference]{ieeeconf}  % Comment this line out if you need a4paper

\usepackage{float}
\usepackage{caption}
\usepackage{subcaption}
\usepackage{graphicx}
\usepackage{xcolor}
\usepackage{hyperref}
\usepackage{soul}
\definecolor{brightgreen}{rgb}{0.0, 0.8, 0.0}

\IEEEoverridecommandlockouts                              % This command is only needed if 
\title{\LARGE \bf
Extensible Hook System for Rendesvouz and Docking of a Cubesat Swarm
}

\author{Carlos J. Pérez-del-Pulgar$^{1}$, Antonio López Palomeque$^{1}$, Jesús Juli$^{1}$ and Matteo Madi$^{2}$% <-this % stops a space
\thanks{*This work was supported by Sirin Orbital Systems AG under contract no. 8.06/5.56.6846 between this company and the University of Malaga. The proposed concept was initially developed for the project, "Swarms of CubeSats for kW-scale Space-Based Solar Power (16U4SBSP)" in the scope of ESA's 2023 Campaign, “Innovative Mission Concepts Enabled by Swarms of CubeSats”}% <-this % stops a space
\thanks{$^{1}$Carlos J. Pérez-del-Pulgar, Antonio López-Palomeque and Jesús Juli are with the Space Robotics Laboratory, Mechatronics and Cyber-physical systems research institute, Universidad de Malaga, 29070 Malaga, Spain.
        {\tt\small carlosperez@uma.es}}%
\thanks{$^{2}$Matteo Madi is with Sirin Orbital Systems AG, Zürich, Switzerland.
        {\tt\small madi@sirin-os.com}}%
}

\begin{document}

\maketitle
\thispagestyle{empty}
\pagestyle{empty}

%%%%%%%%%%%%%%%%%%%%%%%%%%%%%%%%%%%%%%%%%%%%%%%%%%%%%%%%%%%%%%%%%%%%%%%%%%%%%%%%
\begin{abstract}
The use of cubesat swarms is being proposed for different missions where cooperation between satellites is required.
Commonly, the cube swarm requires formation flight and even rendezvous and docking, which are very challenging tasks since they required more energy and the use of advanced guidance, navigation and control techniques.
In this paper, we propose the use of an extensible hook system to mitigate these drawbacks,i.e. it allows to save fuel and reduce the system complexity by including techniques that have been previously demonstrated on Earth.
This system is based on a scissor boom structure, which could reach up to five meters for a 4U dimension, including three degrees of freedom to place the end effector at any pose within the system workspace.
We simulated the dynamic behaviour of a cubesat with the proposed system, demonstrating the required power for a 16U cubesat equipped with one extensible hook system is considered acceptable according to the current state of the art actuators.
\end{abstract}

%%%%%%%%%%%%%%%%%%%%%%%%%%%%%%%%%%%%%%%%%%%%%%%%%%%%%%%%%%%%%%%%%%%%%%%%%%%%%%%%
\section{Introduction}

% Intro al uso de cubesat swarm
Cubesats are miniaturised satellites that can reach a size between 10cm x 10cm x 10cm (1U) and 22.63cm x 22.63cm x 45.4cm (16U) and a mass lower than 25kg for a 16U cubesat. Based on standardisation, their development is commonly fast and cheap.
Initially, they were used for education and research, evolving towards a platform to demonstrate early technologies. 
However, they are increasingly used for commercial applications \cite{golkar2023overview}.

It is common to launch a cubesat alone or in constellations, however, the use of a cubesat swarm arise as a novel way of operating them as a single spacecraft with many advantages, i.e. scalability, robustness, etc. 
However, guidance, navigation and control (GNC) of a cubesat swarm is a very challenging task, requiring the use of novel methods and technologies. 
In the field of Guidance, it requires the use of complex motion planning methods that allow to coordinate the cubesat swarm by generating trajectories to achieve a particular formation in space. Navigation requires to know the 6 Degree of Freedom (6DoF) pose of each satellite by combining different propioceptive and extereoceptive sensors. Finally, control must guarantee the generated trajectory is followed by controlling the actuators.

Currently, cubesat swarm have been proposed for different applications. For example, in \cite{budianu2015swarm}, authors proposed the use of a distributed  cubesat swarm to perform low frequency radio astronomy placing it orbiting the moon. The SWIMSat mission \cite{teja2018orbit} proposed to use a cubesat swarm placed in low earth orbit (LEO) to monitor meteor impacts on the sky above North America. Finally, cubesat swarm has been also proposed in different missions related to Earth observation \cite{liu2022survey}, highlighting the European Space Agency (ESA) Swarm mission to study the dynamics of the Earth magnetic field \cite{van2015precise}.
\begin{figure}
    \centering
    \vspace{0.3cm}
    \includegraphics[width=\columnwidth]{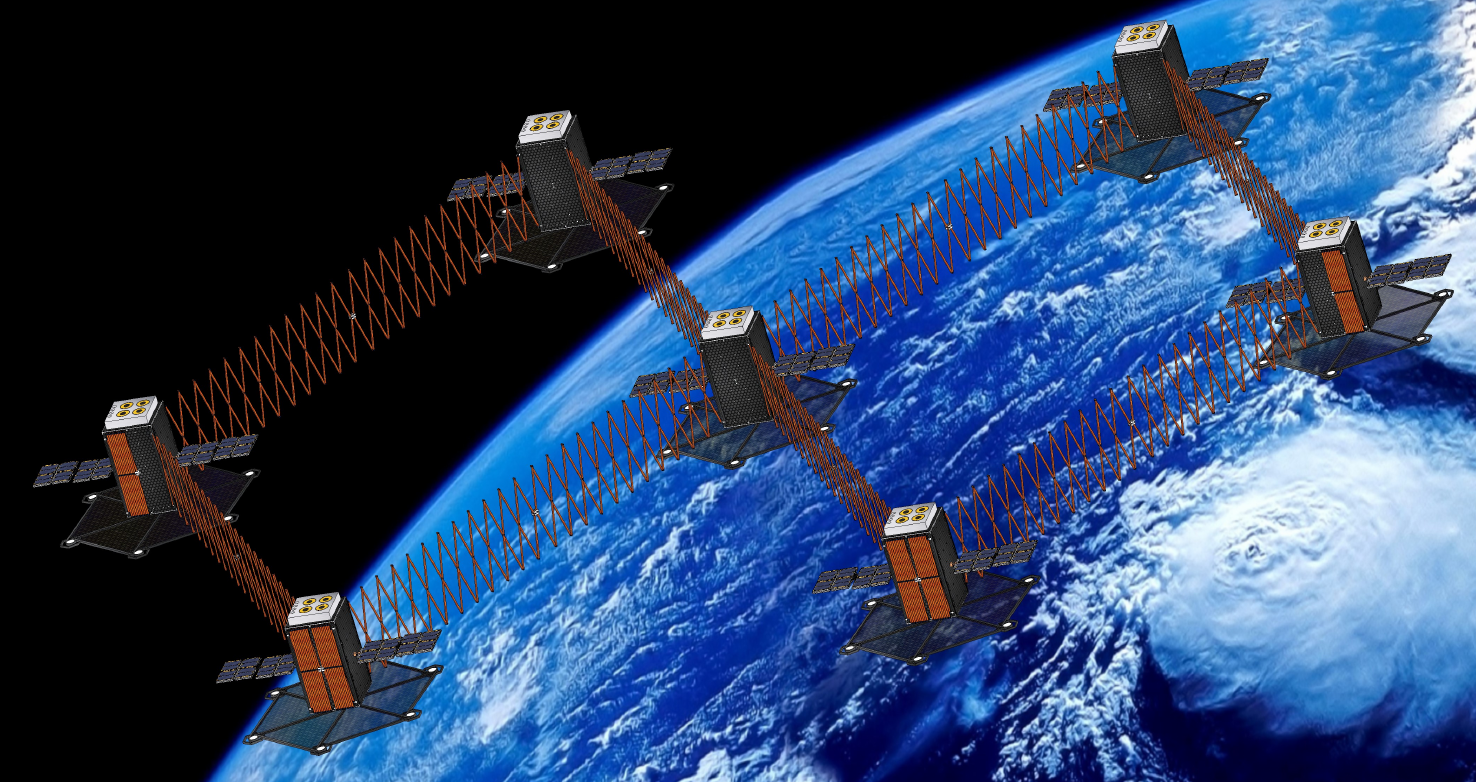}
    \caption{Deployed cubesat swarm with the extensible hook system.}
    \label{fig:swarmcubesats}
\end{figure}

% Necesidad de conexión de cubesat (rendezvouz and docking)
As stated before, cubesat swarm has been investigated for different missions, taking advantage of this configuration to accomplish with their objectives. However, emerging applications in space, such as Active Debris Removal (ADR) \cite{pirat2017mission} and Space-Based Solar Power (SBSP)\cite{curreri2011contemporary} will require orbital Rendezvous and Docking (RVD) operations to capture debris or creating space infrastructures respectively.
In the latter case, RVD can be done by attaching several satellites, e.g. the AAReST mission \cite{underwood2015using}, with the objective of assembling a space telescope based on four 3U cubesats that are docked. 
% Contribuciones del artículo, extensible hook system
Other applications would require to keep the cubesats connected each other without requiring RVD. For example, building large structures in space, such as the proposed one for SBSP \cite{pelton2019space}, which could be made using a cubesat swarm. In these cases, a hook system would be required to connect the cubesat to each other forming the required structure.
Furthermore, an extensible hook system (EHS) provides a big advantage related to avoid the use of thrusters (propellant saving) in two scenarios: when RVD is required, the approaching stage can be done by the EHS, and during a formation flight, where the EHS can be used to keep the cubesat swarm aligned. 

In this sense, this paper proposes an extensible hook system concept based on a scissor boom. This concept has been previously proposed to be used for different applications in CubeSats. For example, to extend a set of sensors 13 cm away from the cubesat \cite{fish2014design}, with the objective of avoiding electronics interference with the sensors, or to deploy solar array systems \cite{trabert2010extendable}.
In this paper, we propose the use of a scissor boom mechanism to connect a cubesat swarm conforming an hexagonal structure as can be seen in Figure \ref{fig:swarmcubesats}.
In order to create this structure, not only the aforementioned mechanism is needed, but also the GNC system to link the cubesats in the optimal way, which will be addressed in this paper by proposing some state of the art methods that could solve the main issues.

\section{Extensible hook system design}
\label{sec:ehs}
\begin{figure}
    %\vspace{0.5cm}
    \centering
    \subfloat[Folded]{
        %\hspace*{-.1in}
        \includegraphics[width=0.41\textwidth]{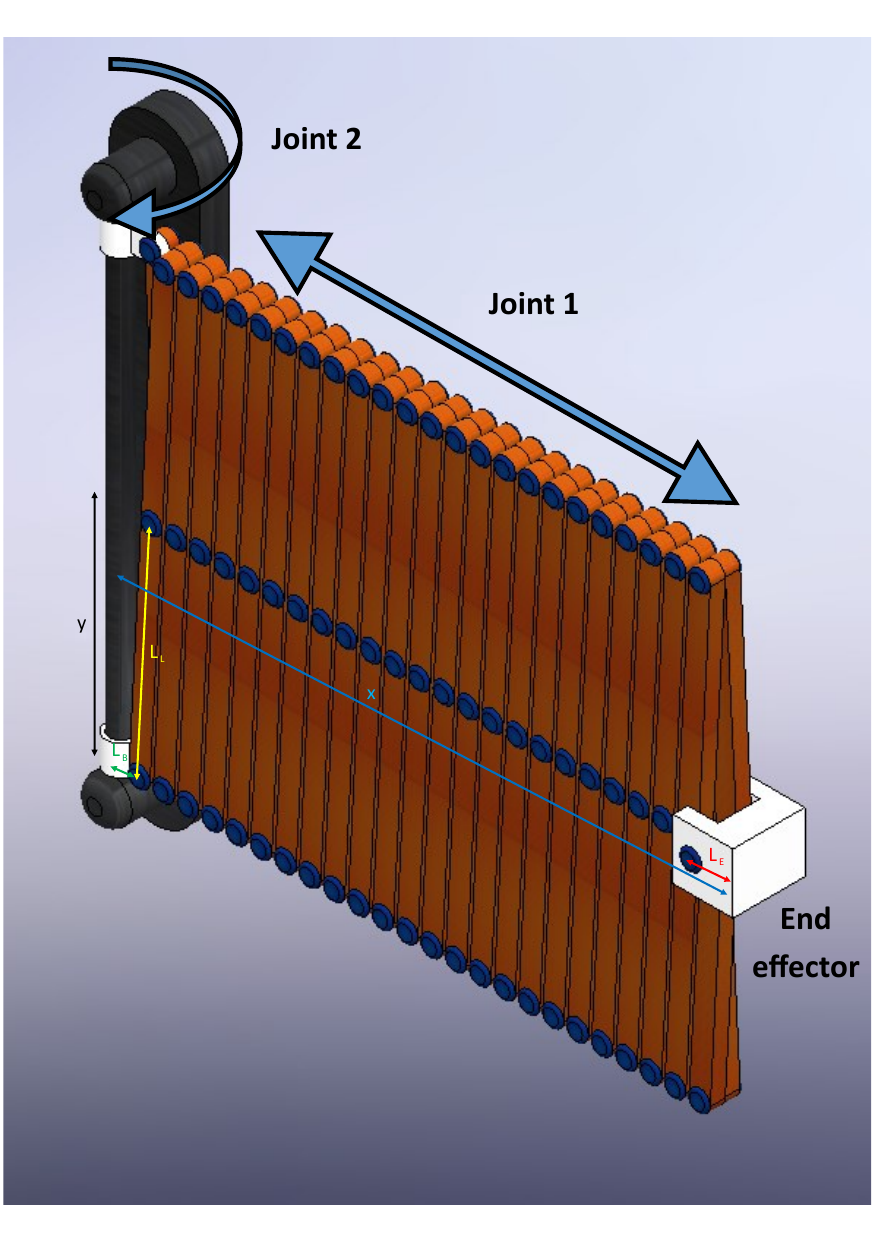}
        % \vphantom{\includegraphics[width=0.31\textwidth,valign=t]{figures/back04.pdf}}
        \label{fig:ehsfolded}
        }
    \\
    \centering
    \subfloat[Deployed (50\%)]{
        %\hspace*{-.1in}
        \includegraphics[width=0.41\textwidth]{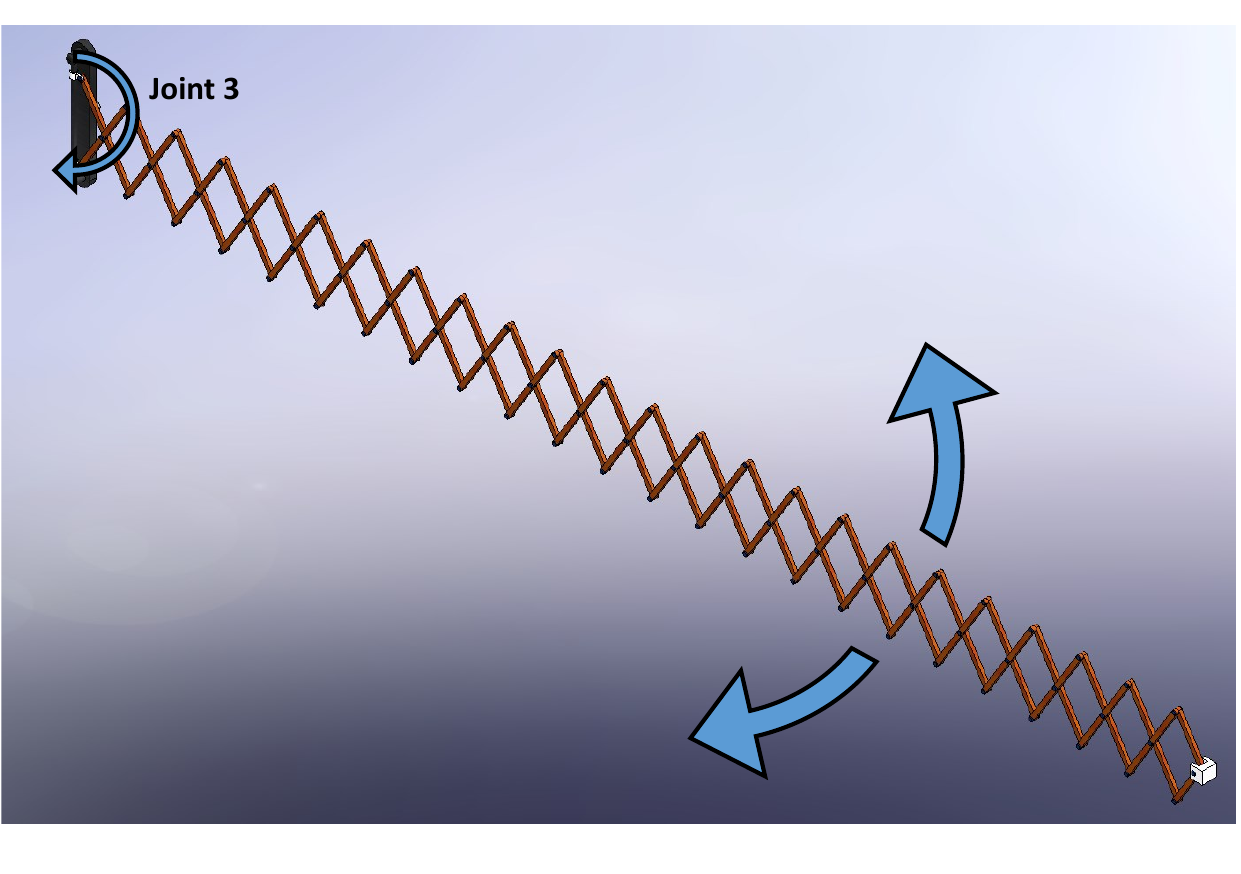}
        % \vphantom{\includegraphics[width=0.31\textwidth,valign=t]{figures/}}
        \label{fig:ehsdeployed}
    }
    \caption{Proposed scissor boom based extensible hook system with 3 joints.}
    \label{fig:scissorboom}
\end{figure}

% Extrensible hook system
The proposed extensible hook system is based on a scissor boom structure with different Degree of Freedoms (DoF) according to the mission requirements and cubesat structure. Figure \ref{fig:scissorboom} shows an initial design of the proposed system. It has been chosen due to its mechanical simplicity, and the capacity of being extended more distance than other structures, such as simple manipulators, etc. However, it is pending of analysing lateral stability according to the used materials and the particular application. The proposed system is initially provided with three joints (3DoF): The first joint allows the system to extend and retract the end effector, actuating on the bar that holds the scissor structure. The second joint allows the main scissor structure to be reorientated. Finally, the whole EHS can be rotated by the third joint. These 3 DoF allow to place the end effector at any position within the EHS workspace.

According to this configuration, equation \ref{eq:ehs} represents the position of the end effector ($x$) w.r.t. the prismatic joint ($y$) on the actuation bar, which depends on the half length of the link ($L_L$), the distance between the actuation bar to the first link ($L_B$), the distance between the last link and the end effector contact point ($L_E$) and the number of pairs of intermediate links ($N$). %El valor de N solo debe tener en cuenta el número de parejas de eslabones intermedios, no se deben contar ni la pareja de eslabones inciales ni la pareja de eslabones finales.
It is worth mentioning this equation includes non-linearities that should be taken into consideration to design the joint controller.

\begin{equation}
\label{eq:ehs}
    x(y) = L_E + L_B + (2N + 3) \left( \sqrt{L_L^2 - y^2} \right)
\end{equation}

\begin{figure}
    \vspace{0.5cm}
    \centering
    \subfloat[Cubesat side with two EHS.]{
        %\hspace*{-.1in}
        \includegraphics[width=0.42\textwidth]{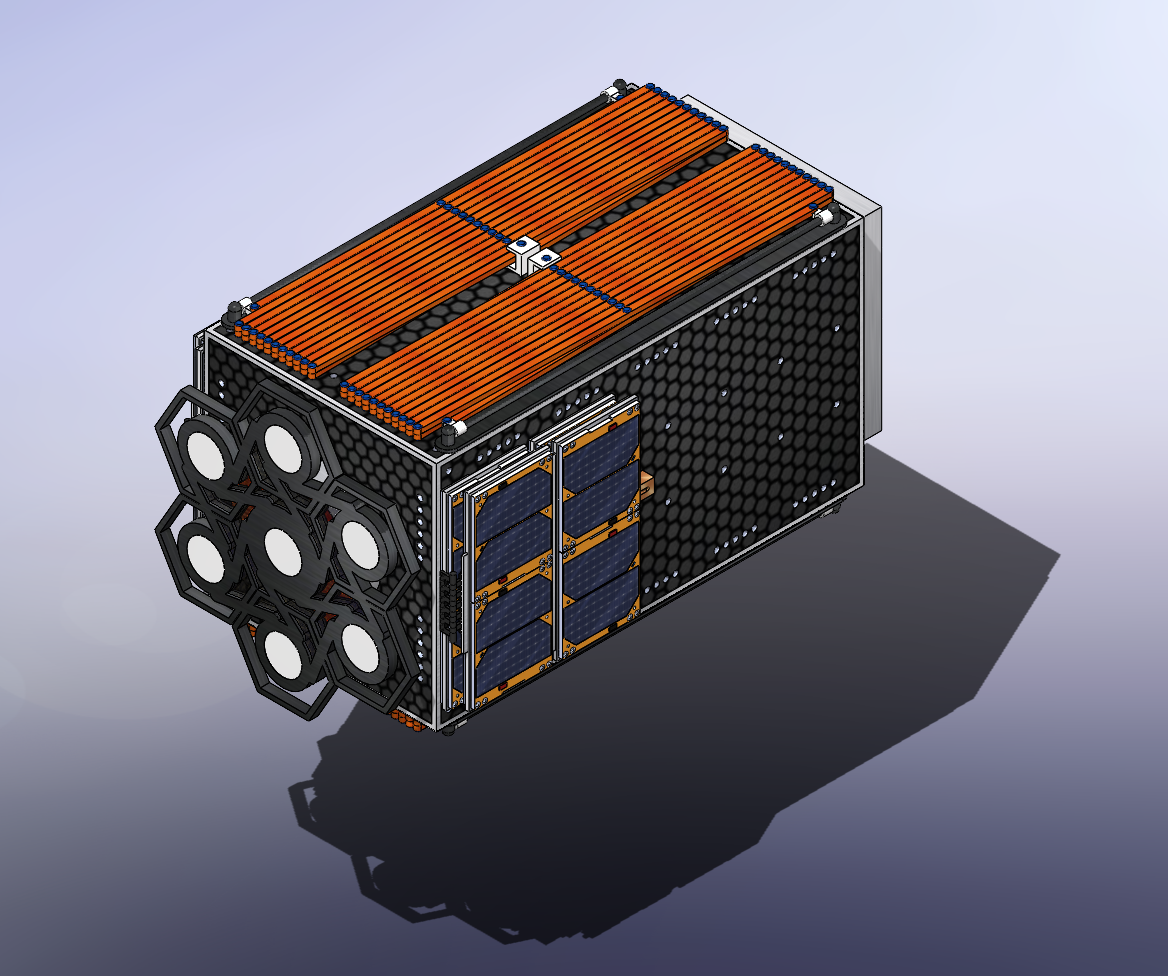}
        % \vphantom{\includegraphics[width=0.31\textwidth,valign=t]{figures/back04.pdf}}
        \label{fig:cubesatSide}
    }
    \\
    \centering
    \subfloat[Cubesat with four EHS deployed.]{
        %\hspace*{-.1in}
        \includegraphics[width=0.42\textwidth]{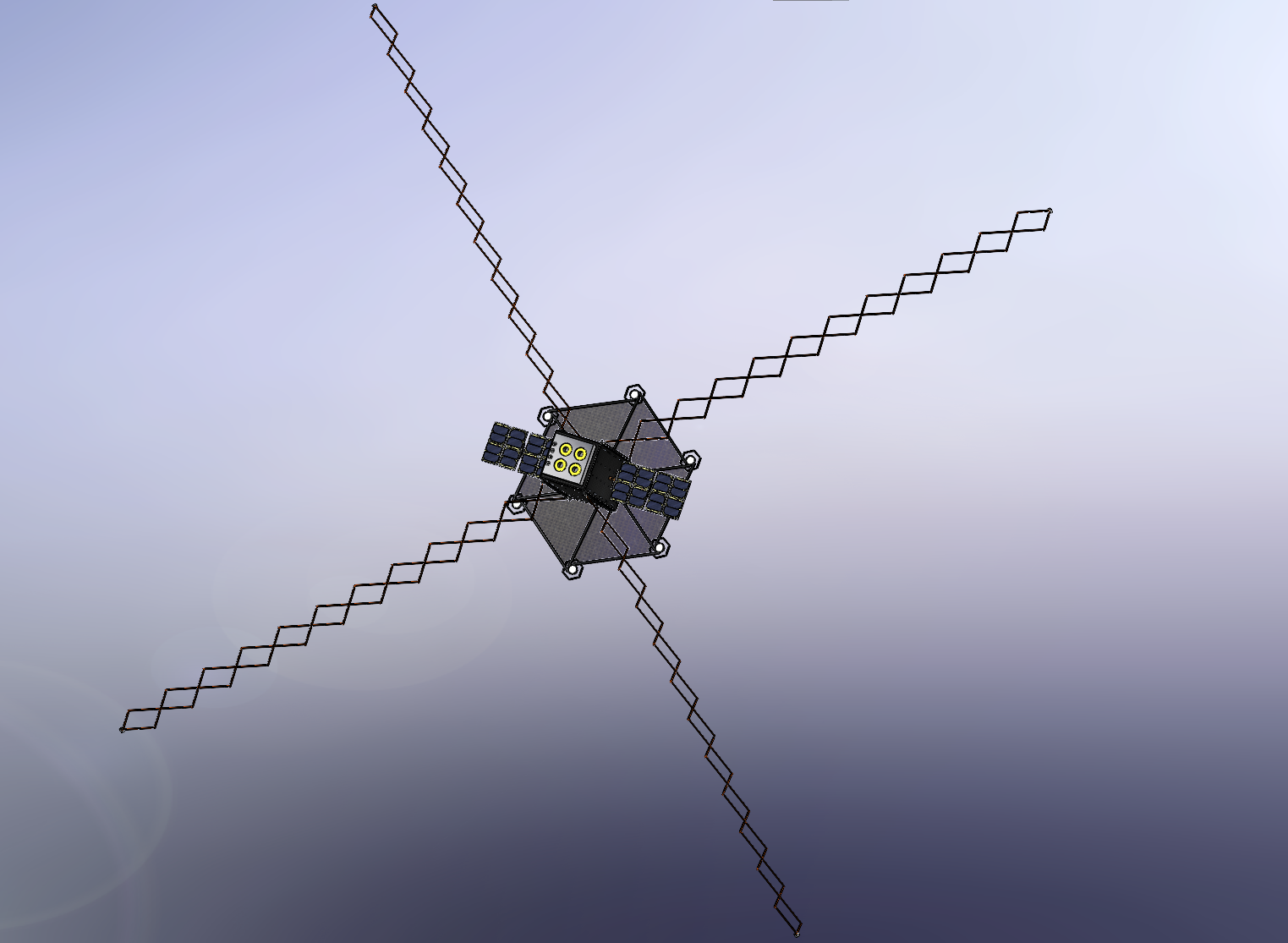}        
        \label{fig:cubesatdeployed}
    }
    \caption{Cubesat configuration with four EHS folded and deployed.}
    \label{fig:cubesat}
\end{figure}

% Deployment on cubesat
Based on this configuration, the EHS has been adapted to be mounted on a cubesat with a 4U total size, i.e. 10cm x 10cm x 45.4cm. This configuration allows the EHS to be extended up to 5m, allowing to place four EHS (two per side) in a 16U cubesat. Figure \ref{fig:cubesatSide} illustrates the already described configuration, in which two EHS are placed on a side.
The EHS could be adapted to different applications by reducing its size, and therefore its reachability. However, in this paper, the main objective of this EHS is to be able to connect the cubesat swarm from a 10m distance, keeping them connected in a formation flight.

The end effector will be based on a configuration with permanent and electro magnets, allowing them to connect EHSs between them, using the electromagnets to increase the power during the docking stage, or to compensate the permanent magnets to undock both cubesats. A clear example of this configuration has been published with application on drones \cite{saldana2018modquad}, where they were equipped with permanent magnets that attracted each other to create a unique structure on air. 
The main drawback of the use of magnetic fields is that it could affect the cubesat sensors, such as the magnetometers, which are commonly used for cubesat navigation.

\section{Guidance, navigation and control}
\begin{figure}
    \centering
    \vspace{0.5cm}
    \includegraphics[width=\columnwidth]{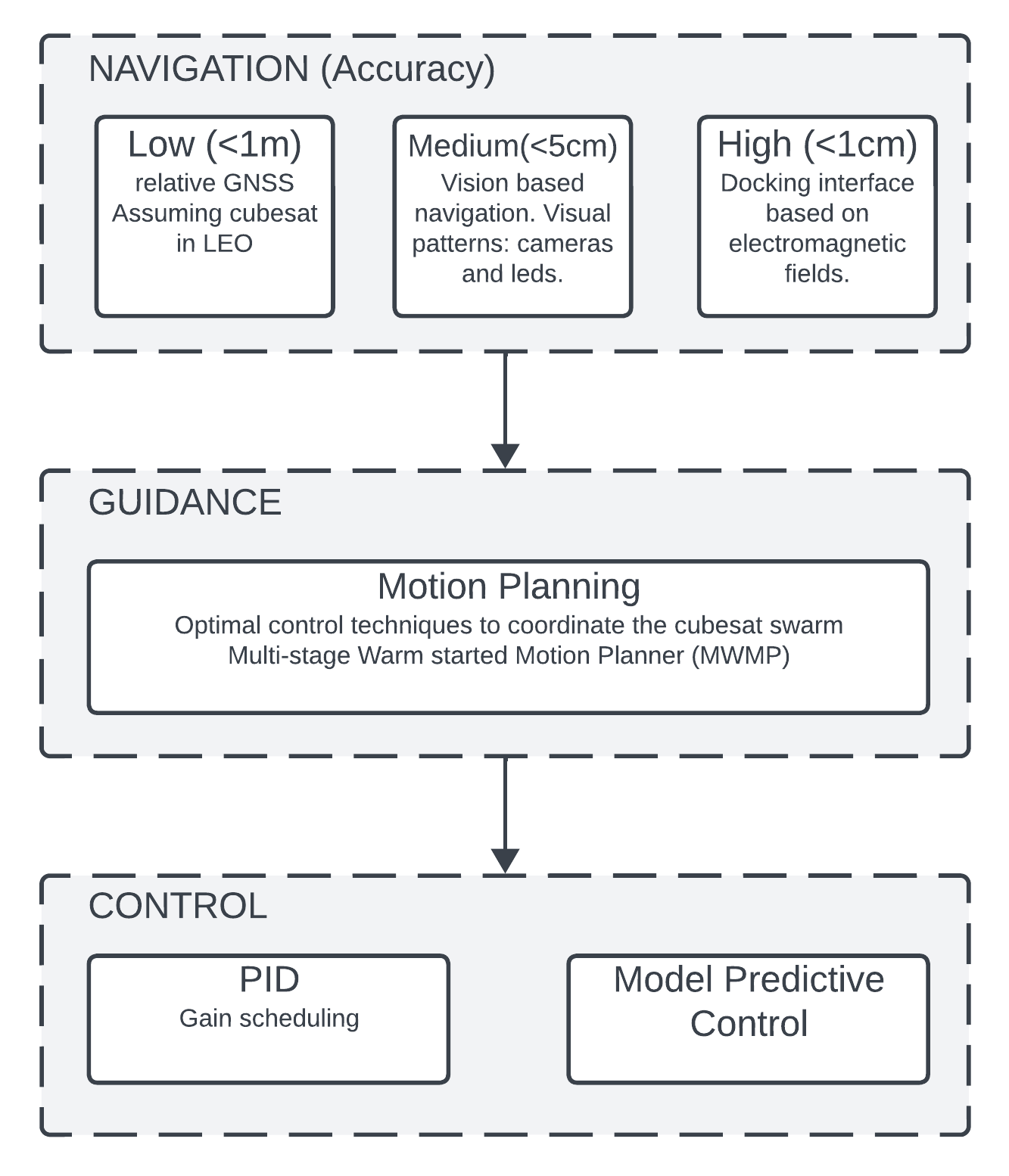}
    \caption{Proposed Guidance, Navigation and Control architecture.}
    \label{fig:gnc}
\end{figure}

To perform cubesat RVD using the EHS, the GNC subsystem needs to be defined for this purpose. The main requirements of it is to guarantee the cubesats can locate each other with low to high accuracy depending on their distance. On the other hand, the guidance component will require to plan the motion for the cubesat swarm to combine their motions in order to be connected through the EHS. Finally, all the cubesat actuators are required to be controlled to follow the defined motion plan by the guidance component.

Figure \ref{fig:gnc} illustrates the proposed GNC subsystem. Firstly, the navigation component has been splitted into three stages, each one requiring lower or higher accuracy. When the cubesats are far ($>5m$), low accuracy is required ($<1m$). 
This accuracy can be reach by using Global Navigation Satellite Services (GNSS), which in the case of relative localisation in low earth orbit (LEO), can reach 0.5m and below \cite{giralo2019distributed}.
Once the cubesats are approaching, the navigation component will require higher accuracy ($<5cm$). For this purpose, our proposal would be to include a vision based navigation system, which would be made up of led patterns and cameras. using this information, localisation of the cubesat swarm could be done with an accuracy in the order of 1cm \cite{pirat2018vision}.
Finally, when the cubesats are about to dock using the EHS, high resolution localisation is needed. In this case, this localisation is replaced by electromagnetic fields on the EHS end effector that will generate the corresponding forces to attract both end effectors by hooking them.

As regards the guidance component, it will require to perform motion planning for the cubesat swarm, including the EHS. Our proposal would be to use a motion planner that allows the swarm to plan all the cubesat motion taking into consideration non-linearities and constraints. 
Multi-staged warm started motion planners arise as a solution to be evaluated. The main advantage of this method is that it allows to compute an optimal plan faster with an increased success calculation \cite{paz2023multi}, i.e. these planners sometimes are not able to find a solution.

Finally, control requires not only commanding the cubesat to follow the predefined motion plan by the guidance component, but also it requires to control all the actuators, including those that belong to the EHS. According to equation \ref{eq:ehs}, the extensible joint (joint 3 in Figure \ref{fig:ehsfolded}) is non-linear, therefore a classical PID controller cannot be used for that purpose. Our proposal would be to evaluate the use of Model Predictive Control (MPC) and gain scheduling. The first one has the advantage of allowing to control the actuator taking into consideration a realistic model, therefore predicting its behaviour, however it requires more computation. On the other hand, gain scheduling does not require a lot of computation, but the control is simpler.

\section{Simulation results}
\label{sec:simulation}
% Que se pretende demostrar con los experimentos en simulación
The design of the EHS requires to be validated through some simulations. They will provide meaninful information about the required forces, torques an energy to move both the cubesat and the EHS, allowing them to chose the best actuators, i.e. thrusters and motors.
The parameters that were used for the cubesat and the EHS are shown in Table \ref{tab:parameters}.

\begin{table}[]
    \centering
    \vspace{0.4cm}
    \begin{tabular}{|c|c|}
    \hline
         \textbf{Parameter} &  \textbf{Value} \\ \hline
         Total mass (w/o EHS) & 27.7 kg \\ \hline
         EHS mass & 2.41 kg \\ \hline
         Cubesat size & 246.3 x 246.3 x 454 mm (16U) \\ \hline
         EHS size (folded) & 122.51 x 444.54 x 200 mm (4U) \\ \hline
         EHS max. distance & 5,026 mm \\ \hline
         EHS links & 24 \\ \hline
         EHS joint friction & 0.01 N s/m \\ \hline 
         Cubesat longitudinal max. force & 0.1 N \\ \hline 
         Cubesat max. torque & 0.2 Nm \\ \hline 
         EHS link max. force & 20 N \\ \hline 
         EHS joint max. torque & 200 mNm \\ \hline
    \end{tabular}
    \caption{Cubesat parameters during simulation.}
    \label{tab:parameters}
\end{table}

The actuators were controlled by adjusting a proportional, integrative and derivative (PID) controller, and a trapezoidal trajectory generator was included in the model to perform the simulations.

% Herramientas empleadas
Simulations were made using Matlab Simscape Multibody. This tool allows us to assemble all the pieces, which were previously designed using SolidWorks, and simulate their dynamics by solving the whole mechanic system equations.
Moreover, this tool allows us to generate animations of the system and it can be used to test different controllers by using the hardware and software in the loop concepts.

% Sistema de GNC
During the simulation Guidance and Navigation was omitted, due to their complexity they are proposed as future work. Navigation was replaced by the abosolute localisation that provides the simulator, and control was implemented using PIDs and gain scheduling for the case of the joint 3 of the EHS.

% Experimentos realizados
To validate the proposed system, two simulations were carried out. The first one related to move the cubesat from an initial pose to a final one, and the second simulation was related to the extension of the EHS and the required forces and torques to perform the motion.

\subsection{Cubesat trajectory tracking}
% Poner resultados de mover el cubesat 50m y que alcance una posición y orientación concreta. Incluir gráficas: fuerzas y pares necesarios para mover el cubesat. Valor total en N*s para cada eje.
\begin{figure}
    %\vspace{0.5cm}
    \centering
    \subfloat[Forces]{
        %\hspace*{-.1in}
        \includegraphics[width=0.5\textwidth]{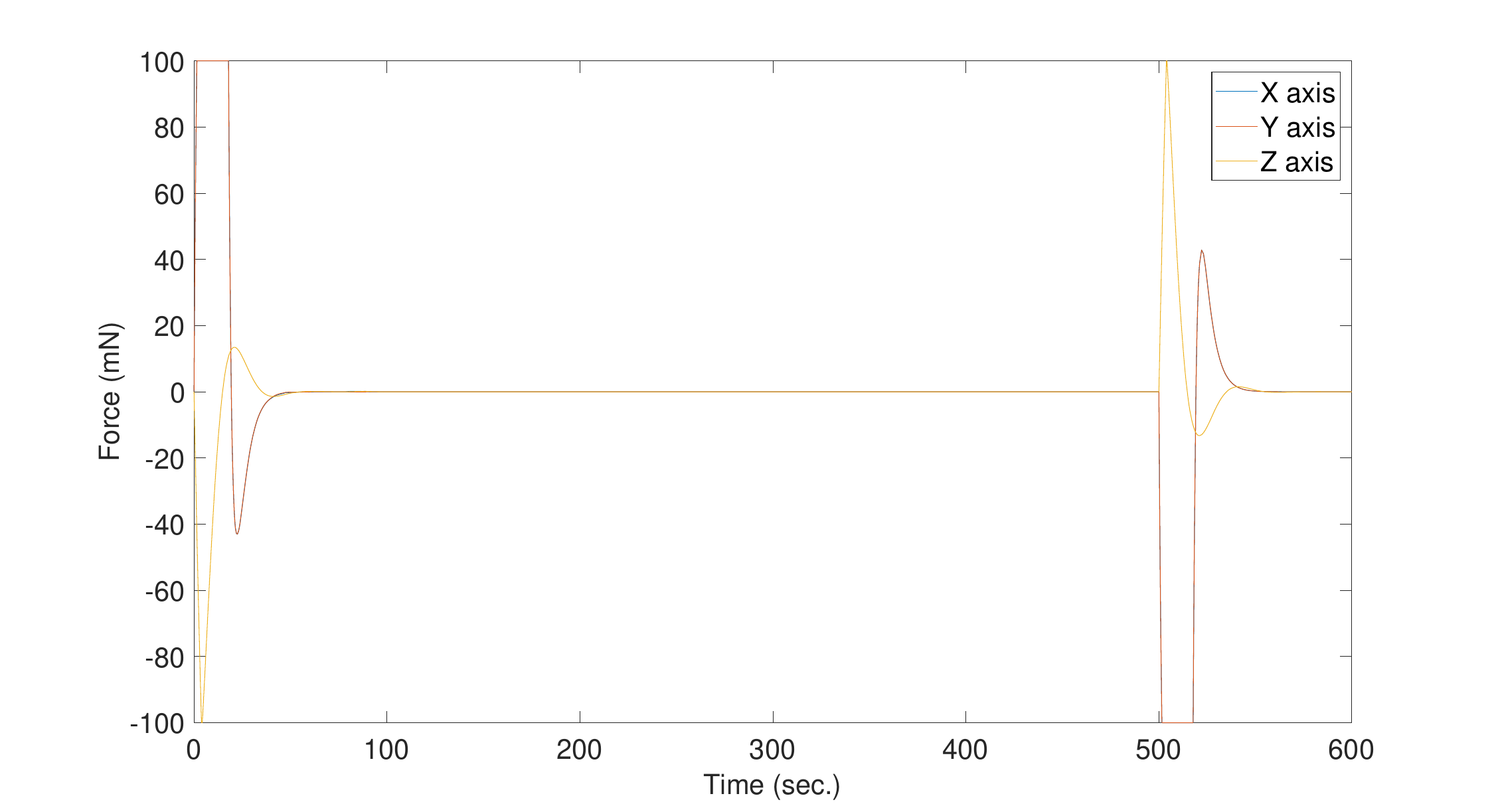}        
        \label{fig:cubesatForcesTrajectory}
    }
    \\
    \centering
    \subfloat[Torques]{
        %\hspace*{-.1in}
        \includegraphics[width=0.5\textwidth]{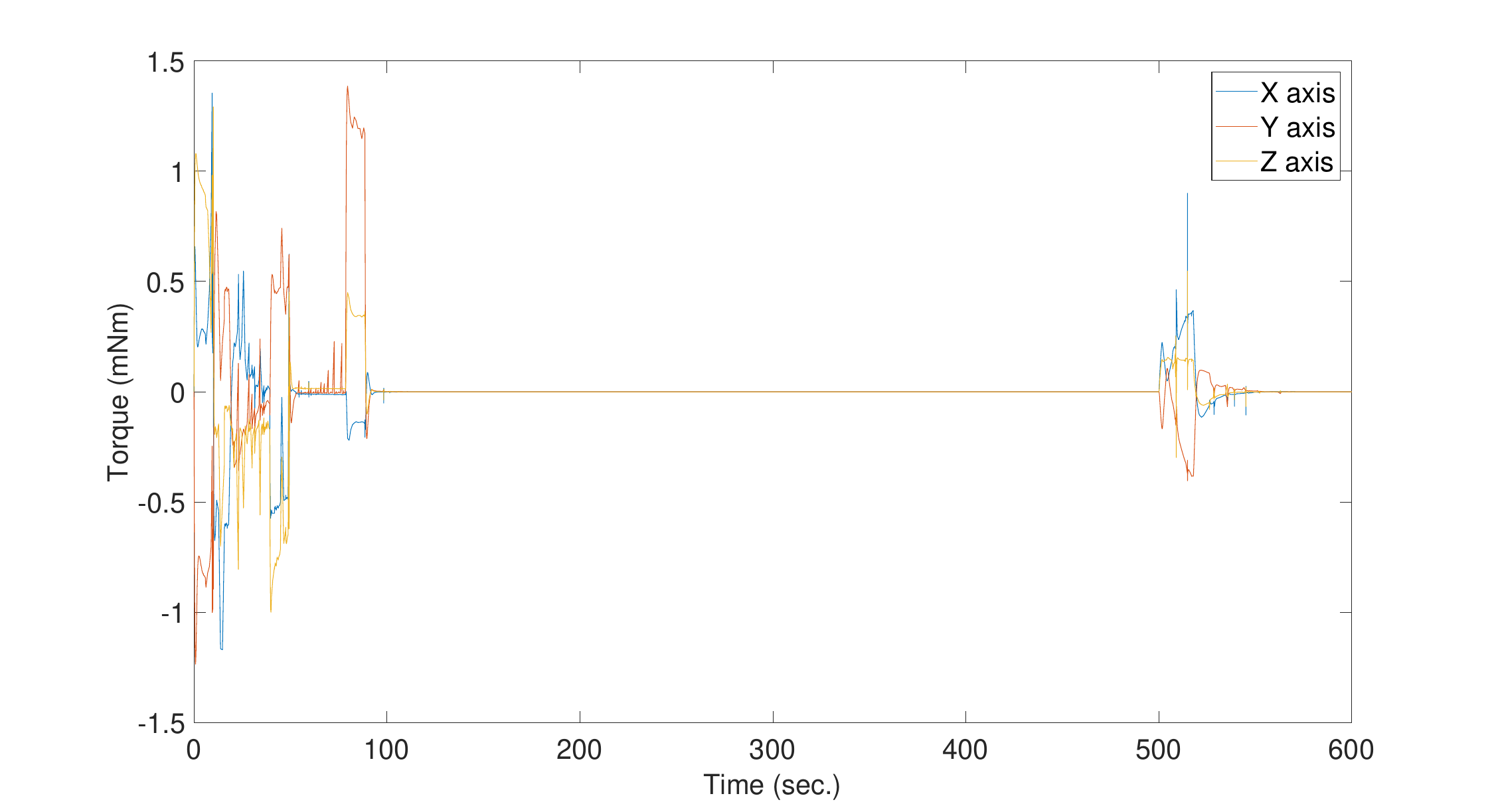}
        % \vphantom{\includegraphics[width=0.31\textwidth,valign=t]{figures/back04.pdf}}
        \label{fig:cubesatTorquesTrajectory}
    }
    
    \caption{Required torques and forces to move the cubesat 36.7523 m in 500 s following a trapezoidal profile.}
    \label{fig:cubesatTorquesForcesTrajectory}
\end{figure}
During this simulation, the objective was to measure the required forces and energy to perform a manoeuvre that would place the cubesat on a particular pose. 
The simulation was configured to move the cubesat from $(0,0,0,0,0,0)$ to $(25m,25m,-10m,\pi/8rad,-\pi/4rad,\pi/8rad)$ in $(X,Y,Z,\alpha,\beta,\gamma)$ position and orientation coordinates, traversing a distance of 36.7523 m in 500 seconds.
Figure \ref{fig:cubesatTorquesForcesTrajectory} shows the obtained results, where the displacement required a maximum longitudinal forces 100 mN on the Y axis, and a maximum torque 1.385 for a rotation on the Y axis.
Moreover, the motion required a total energy of $(4.3296Ns, 4.3248Ns, 1.7377Ns)$  for X,Y,Z.

\subsection{Extensible Hook System Deployment}
% Gráfica con par de rotación, fuerza del prismatic joints. Otra gráfica con fuerzas y pares del cubesat para compensar.
% Comentar los pares y fuerzas máximos necesarios para mover el sistema y compensar sus movmientos. Analizar también la energía necesaria por parte de los propulsores.

\begin{figure}
    %\vspace{0.5cm}
    \centering
    \subfloat[Forces]{
        %\hspace*{-.1in}
        \includegraphics[width=0.5\textwidth]{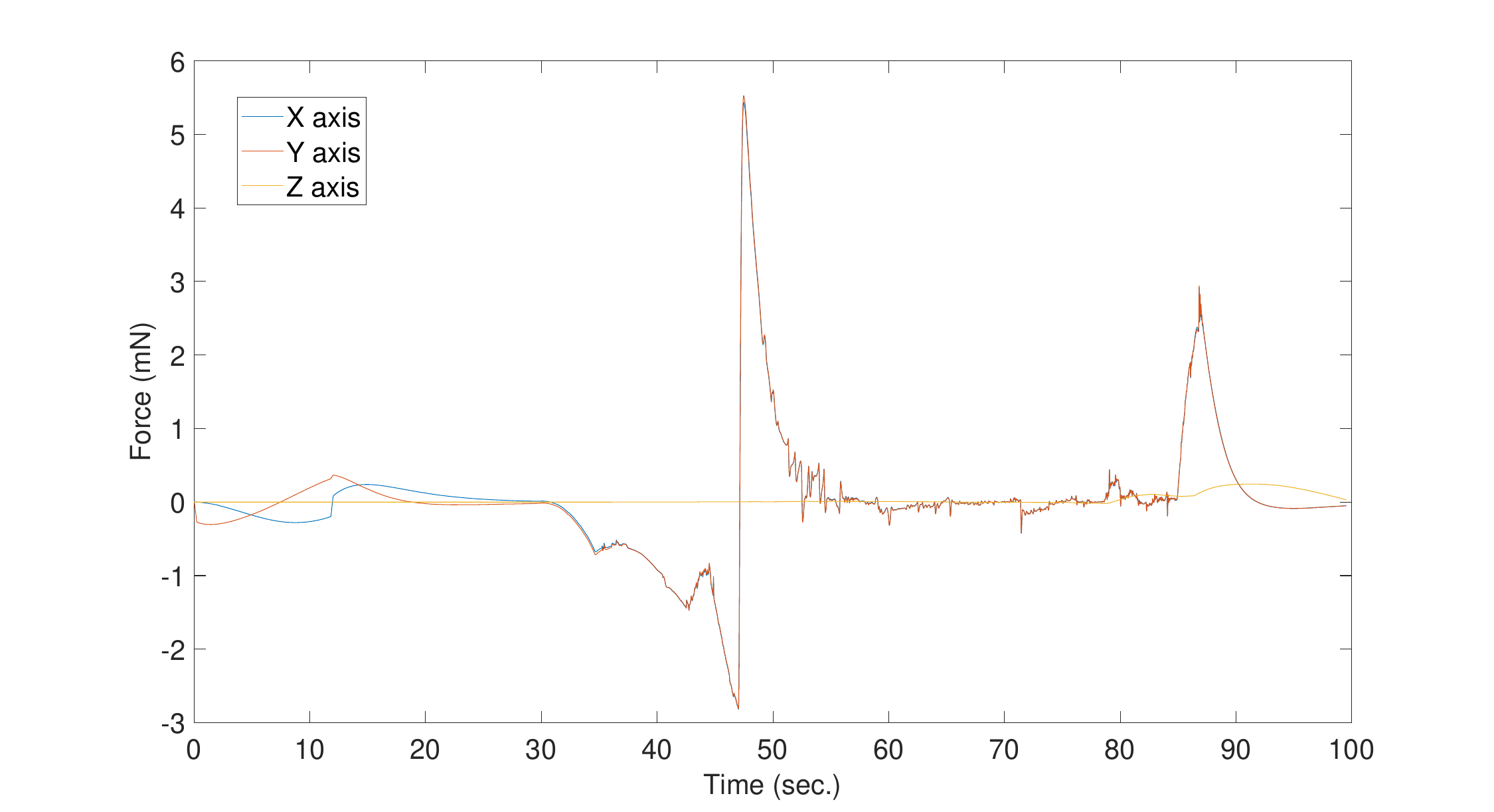}        
        \label{fig:cubesatForces}
    }    
    \\
    \centering
    \subfloat[Torques]{
        %\hspace*{-.1in}
        \includegraphics[width=0.5\textwidth]{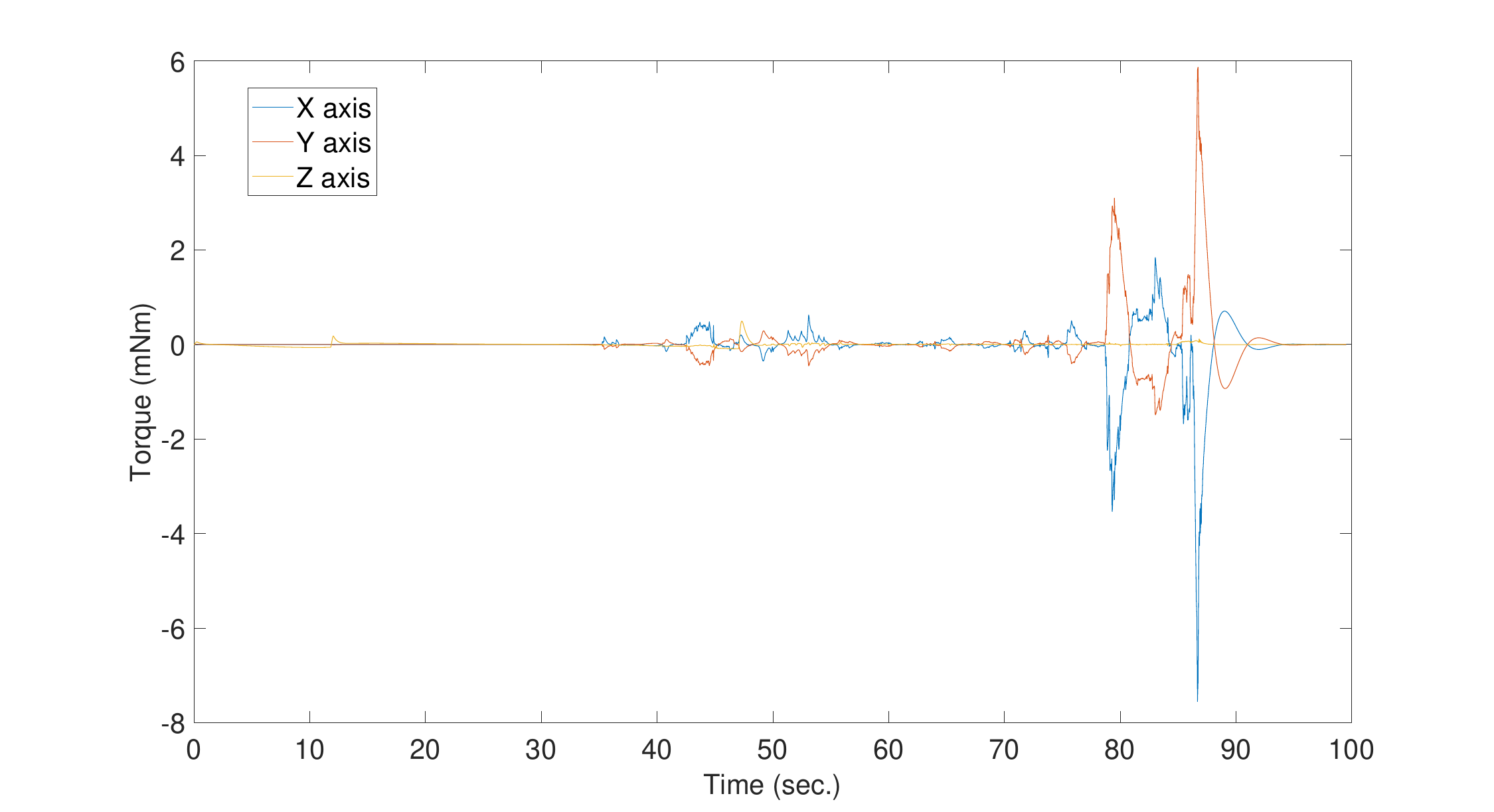}
        % \vphantom{\includegraphics[width=0.31\textwidth,valign=t]{figures/back04.pdf}}
        \label{fig:cubesatTorques}
    }
    \caption{Required torques and forces to keep the Cubesat on a fixed pose during the motion of the EHS.}
    \label{fig:cubesatTorquesForces}
\end{figure}

\begin{figure}
    \centering
    \vspace{0.5cm}
    \includegraphics[width=\columnwidth]{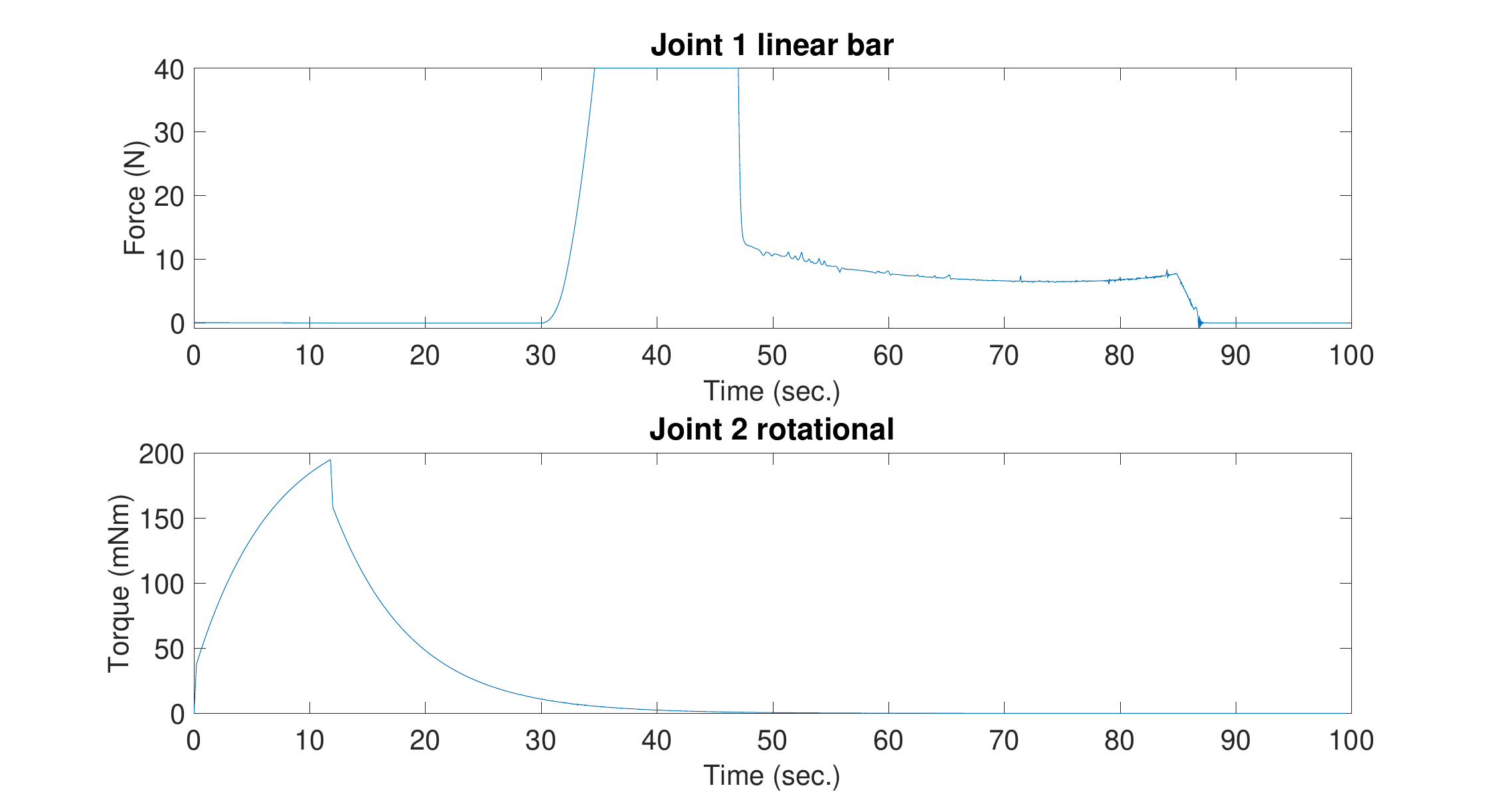}
    \caption{Required Force and Torque to rotate and deploy the EHS.}
    \label{fig:ehsforcestorques}
\end{figure}
To demonstrate the required forces to move the EHS, the cubesat was controlled to keep the initial pose while the EHS was rotated 135 degres and deployed 4.5m.
Figure \ref{fig:cubesatTorquesForces} shows the resulting force and torque to keep the cubesat with a fixed pose. As can be seen, the maximum required forces were 5.527 mN for the X and Y axes, and 2.942 mN for the Z axis. The system required a total energy of $(40mNs, 6.8mNs,39.9mNs)$  for the X, Y and Z axes respectively.  
The maximum required torque was $(-7.553mNm, 5.806mNm,0.1389mNm)$ for the X, Y and Z axes.
On the other hand, the actuator of the joint 1 from the EHS was limited to 40N, and the joint 2 required a maximum torque 194.1 mNm.

To illustrate both simulations, a video\footnote{\url{https://youtu.be/An_atSlO0tM}} with the animations was recorded. It shows how the cubesat was moved to another location and the EHS was deployed.

\section{Conclusions}
\label{sec:conclusions}
% Proposed EHS design
This paper proposed an extensible hook system that allows a cubesat swarm to be connected with the main objective of saving thruster propellant. This propellant saving could be achieved in two stages: during the approaching stage, by bringing the cubesat closer, and the formation flight, keeping the cubesat swarm together by the use of the EHS that connects each other.

% GNC
However, the use of the EHS in combination with the cubesat actuators requires an advanced Guidance, Navigation and Control architecture. This paper proposed a three level navigation component by combining GNSS, vision based navigation and the use of electromagnets for the docking stage. In the case of guidance, it requires complex motion planning methods to coordinate the cubesat swarm. Finally, control requires classical PIDs in combination with other advanced techniques, such as gain scheduling and Model Predictive Control due to the non-linearities in the EHS.

% Simulation results
The proposed system was simulated using Matlab Simscape Multibody, obtaining meaningful information about the required forces and torques to move the cubesat and the EHS. Obtained measurements are realistic, i.e. they can be generated by thrusters, reaction wheels and motors. It is pending of selecting the particular actuators according to the obtained measurements.

% Future work
The proposed concept has been desgined and evaluated from a mechanical and dynamic point of view, however, it is pending of investigating and developing the guidance, navigation and control stack that would allow a cubesat swarm to be connected through the proposed system.
This challenge is proposed as future work, with the aim of using the current simulation environment to test the GNC stack using the software in the loop concept.

%%%%%%%%%%%%%%%%%%%%%%%%%%%%%%%%%%%%%%%%%%%%%%%%%%%%%%%%%%%%%%%%%%%%%%%%%%%%%%%%%%%%%%%%%%%%%
\addtolength{\textheight}{-13cm}   % This command serves to balance the column lengths
                                  % on the last page of the document manually. It shortens
                                  % the textheight of the last page by a suitable amount.
                                  % This command does not take effect until the next page
                                  % so it should come on the page before the last. Make
                                  % sure that you do not shorten the textheight too much.

\bibliographystyle{ieeetr}
\bibliography{bibliography}

\begin{thebibliography}{10}

\bibitem{golkar2023overview}
A.~Golkar, ``Overview of the new space cubesat market,'' in {\em Next Generation CubeSats and SmallSats}, pp.~605--620, Elsevier, 2023.

\bibitem{budianu2015swarm}
A.~Budianu, A.~Meijerink, and M.~J. Bentum, ``Swarm-to-earth communication in olfar,'' {\em Acta astronautica}, vol.~107, pp.~14--19, 2015.

\bibitem{teja2018orbit}
R.~teja Nallapu, H.~Kalita, and J.~Thangavelautham, ``On-orbit meteor impact monitoring using cubesat swarms,'' in {\em The Advanced Maui Optical and Space Surveillance Technologies Conference}, p.~11, 2018.

\bibitem{liu2022survey}
S.~Liu, P.~I. Theoharis, R.~Raad, F.~Tubbal, A.~Theoharis, S.~Iranmanesh, S.~Abulgasem, M.~U.~A. Khan, and L.~Matekovits, ``A survey on cubesat missions and their antenna designs,'' {\em Electronics}, vol.~11, no.~13, p.~2021, 2022.

\bibitem{van2015precise}
J.~van~den IJssel, J.~Encarna{\c{c}}{\~a}o, E.~Doornbos, and P.~Visser, ``Precise science orbits for the swarm satellite constellation,'' {\em Advances in Space Research}, vol.~56, no.~6, pp.~1042--1055, 2015.

\bibitem{pirat2017mission}
C.~Pirat, M.~Richard-Noca, C.~Paccolat, F.~Belloni, R.~Wiesendanger, D.~Courtney, R.~Walker, and V.~Gass, ``Mission design and gnc for in-orbit demonstration of active debris removal technologies with cubesats,'' {\em Acta Astronautica}, vol.~130, pp.~114--127, 2017.

\bibitem{curreri2011contemporary}
P.~A. Curreri and M.~K. Detweiler, ``A contemporary analysis of the o'neill-glaser model for space-based solar power and habitat construction,'' tech. rep., 2011.

\bibitem{underwood2015using}
C.~Underwood, S.~Pellegrino, V.~J. Lappas, C.~P. Bridges, and J.~Baker, ``Using cubesat/micro-satellite technology to demonstrate the autonomous assembly of a reconfigurable space telescope (aarest),'' {\em Acta Astronautica}, vol.~114, pp.~112--122, 2015.

\bibitem{pelton2019space}
J.~N. Pelton and J.~N. Pelton, ``Space-based solar power satellite systems,'' {\em Space 2.0: Revolutionary Advances in the Space Industry}, pp.~103--114, 2019.

\bibitem{fish2014design}
C.~Fish, C.~Swenson, G.~Crowley, A.~Barjatya, T.~Neilsen, J.~Gunther, I.~Azeem, M.~Pilinski, R.~Wilder, D.~Allen, {\em et~al.}, ``Design, development, implementation, and on-orbit performance of the dynamic ionosphere cubesat experiment mission,'' {\em Space Science Reviews}, vol.~181, pp.~61--120, 2014.

\bibitem{trabert2010extendable}
R.~Trabert, A.~Klesh, P.~Senatore, P.~Martinchek, D.~Becker, A.~Chou, C.~Hoffman, C.~Hoffman, N.~McKay, J.~Nash, {\em et~al.}, ``The extendable solar array system: a modular nanosatellite power system,'' in {\em AIAA/AAS Astrodynamics Specialist Conference}, p.~7654, 2010.

\bibitem{saldana2018modquad}
D.~Saldana, B.~Gabrich, G.~Li, M.~Yim, and V.~Kumar, ``Modquad: The flying modular structure that self-assembles in midair,'' in {\em 2018 IEEE International Conference on Robotics and Automation (ICRA)}, pp.~691--698, IEEE, 2018.

\bibitem{giralo2019distributed}
V.~Giralo and S.~D’Amico, ``Distributed multi-gnss timing and localization for nanosatellites,'' {\em Navigation}, vol.~66, no.~4, pp.~729--746, 2019.

\bibitem{pirat2018vision}
C.~Pirat, F.~Ankersen, R.~Walker, and V.~Gass, ``Vision based navigation for autonomous cooperative docking of cubesats,'' {\em Acta Astronautica}, vol.~146, pp.~418--434, 2018.

\bibitem{paz2023multi}
G.~J. Paz-Delgado, C.~J. P{\'e}rez-del Pulgar, M.~Azkarate, F.~Kirchner, and A.~Garc{\'\i}a-Cerezo, ``Multi-stage warm started optimal motion planning for over-actuated mobile platforms,'' {\em Intelligent Service Robotics}, pp.~1--17, 2023.

\end{thebibliography}

\end{document}